\pdfoutput = 1
\documentclass{article}
\usepackage{times}
\usepackage[comma,authoryear]{natbib}
\usepackage[english]{babel}
\usepackage{blindtext}
\usepackage{graphicx}
\usepackage{amsmath, amsthm, amssymb, bbm, bm}
\usepackage{enumerate}
\usepackage{float}      % figure inside minipage,
\usepackage{subcaption}  % does not go well with subfig.
\usepackage{wrapfig}
\usepackage[margin=0cm]{caption}
\usepackage[titletoc, toc]{appendix}
\usepackage{tabularx}

\usepackage{multirow}
\usepackage{makecell}
\usepackage{placeins}  % stop floating

\usepackage[x11names, usenames, dvipsnames, svgnames, table]{xcolor}
\definecolor{firebrick}{rgb}{.698,.133,.133}
\definecolor{mybluelight}{rgb}{0.9, 0.9, 1.}

\usepackage[utf8]{inputenc} % allow utf-8 input
\usepackage[T1]{fontenc}    % use 8-bit T1 fonts
\usepackage{url}            % simple URL typesetting
\usepackage{booktabs, colortbl}       % professional-quality tables
\usepackage{amsfonts}       % blackboard math symbols
\usepackage{nicefrac}       % compact symbols for 1/2, etc.
\usepackage{microtype}      % microtypography

\usepackage{csquotes}
\usepackage{latexsym}

\usepackage[boxruled, vlined, linesnumbered]{algorithm2e}
\SetAlFnt{\small}
\SetAlCapFnt{\small}
\SetAlCapNameFnt{\small}
\usepackage{algorithmic}
\algsetup{linenosize=\tiny}

\let\oldnl\nl% Store \nl in \oldnl
\newcommand{\nonl}{\renewcommand{\nl}{\let\nl\oldnl}}% Remove line number for one line

\usepackage{paralist}

\usepackage{xspace}
\usepackage{soul}
\usepackage{dsfont}
\usepackage{stmaryrd}
\usepackage[textwidth=15mm]{todonotes}
\usepackage{dirtytalk}
\usepackage{pbox}
\usepackage{cprotect}

\usepackage{verbatim}
\usepackage{textcomp}
\usepackage[normalem]{ulem}

\usepackage{mathtools}
\usepackage{etextools}
\usepackage[inline]{enumitem}

\usepackage[colorlinks=true,allcolors=firebrick,bookmarks=false]{hyperref}

\let\OLDthebibliography\thebibliography
\renewcommand\thebibliography[1]{
  \OLDthebibliography{#1}
  \setlength{\parskip}{0pt}
  \setlength{\itemsep}{0pt plus 0.3ex}
}
\newcommand{\abs}[1]{\ensuremath \left| #1 \right|}

\newcommand{\la}{\ \leftarrow\ }

\theoremstyle{definition}

\DeclarePairedDelimiterX{\divx}[2]{(}{)}{%
  #1\;\delimsize\|\;#2%
}

\newcommand*{\ie}{\emph{i.e.}\@\xspace}

\usepackage{style}

\title{Deep Active Learning for Joint Classification \& \\ Segmentation with Weak Annotator
\thanks{Code:~\href{https://github.com/sbelharbi/deep-active-learning-for-joint-classification-and-segmentation-with-weak-annotator}{https://github.com/sbelharbi/deep-active-learning-for-joint-classification-and-segmentation-with-weak-annotator}}}

\renewcommand\footnotemark{}

\renewcommand\footnotemark{}

\author{Soufiane Belharbi$^{1}$, Ismail Ben Ayed$^{1}$, Luke McCaffrey$^{2}$, Eric Granger$^{1}$\\[0.03in]
$^{1}$ LIVIA, Dept. of Systems Engineering, École de technologie supérieure, Montreal, Canada\\
$^{2}$ Goodman Cancer Research Centre, Dept. of Oncology, McGill University, Montreal, Canada\\ [0.05in]
{\footnotesize
 \textcolor{firebrick}{soufiane.belharbi.1@ens.etsmtl.ca, \{ismail.benayed, eric.granger\}@etsmtl.ca, luke.mccaffrey@mcgill.ca}
 }
}

\newcommand{\ignore}[1]{}

\begin{document}

\maketitle

\begin{abstract}
CNN visualization and interpretation methods, like class-activation maps (CAMs), are typically used to highlight the image regions linked to class predictions. These models allow to simultaneously classify images and extract class-dependent saliency maps, without the need for costly pixel-level annotations. However, they typically yield segmentations with high false-positive rates and, therefore, coarse visualisations, more so when processing challenging images, as encountered in histology. To mitigate this issue, we propose an active learning (AL) framework, which progressively integrates pixel-level annotations during training. Given training data with global image-level labels, our deep weakly-supervised learning model jointly performs supervised image-level classification and active learning for segmentation, integrating pixel annotations by an oracle.  Unlike standard AL methods that focus on sample selection, we also leverage large numbers of unlabeled images via pseudo-segmentations (i.e., self-learning at the pixel level), and integrate them with the oracle-annotated samples during training. We report extensive experiments over two challenging benchmarks -- high-resolution medical images (histology GlaS data for colon cancer) and natural images (CUB-200-2011 for bird species). Our results indicate that, by simply using random sample selection, the proposed approach can significantly outperform state-of the-art CAMs and AL methods, with an identical oracle-supervision budget. Our code is publicly available.
\end{abstract}

% ===================================================================================================
%
%                                      INTRODUCTION
%
% ===================================================================================================
\section{Introduction}
\label{sec:introduction}

Image classification and segmentation are fundamental tasks in many visual recognition applications involving natural and medical images. Given a large image dataset annotated with global image-level labels for classification or with pixel-level labels for segmentation, deep learning (DL) models achieve state-of-the-art performances for these tasks  \citep{dolz20183d,Goodfellow-et-al-2016,krizhevsky12,LITJENS201760,LongSDcvpr15,Ronneberger-unet-2015}. However, the impressive accuracy of such fully-supervised learning models comes at the expense of a considerable cost for collecting and annotating large image data sets. While the acquisition of global image-level annotation can be relatively inexpensive, pixel-wise annotation involves a laborious process, a difficulty further accrued by the requirement of domain expertise, as in medical imaging, which increases the annotation costs.
\begin{figure}[t!]
\centering
  \centering
  \includegraphics[scale=0.49]{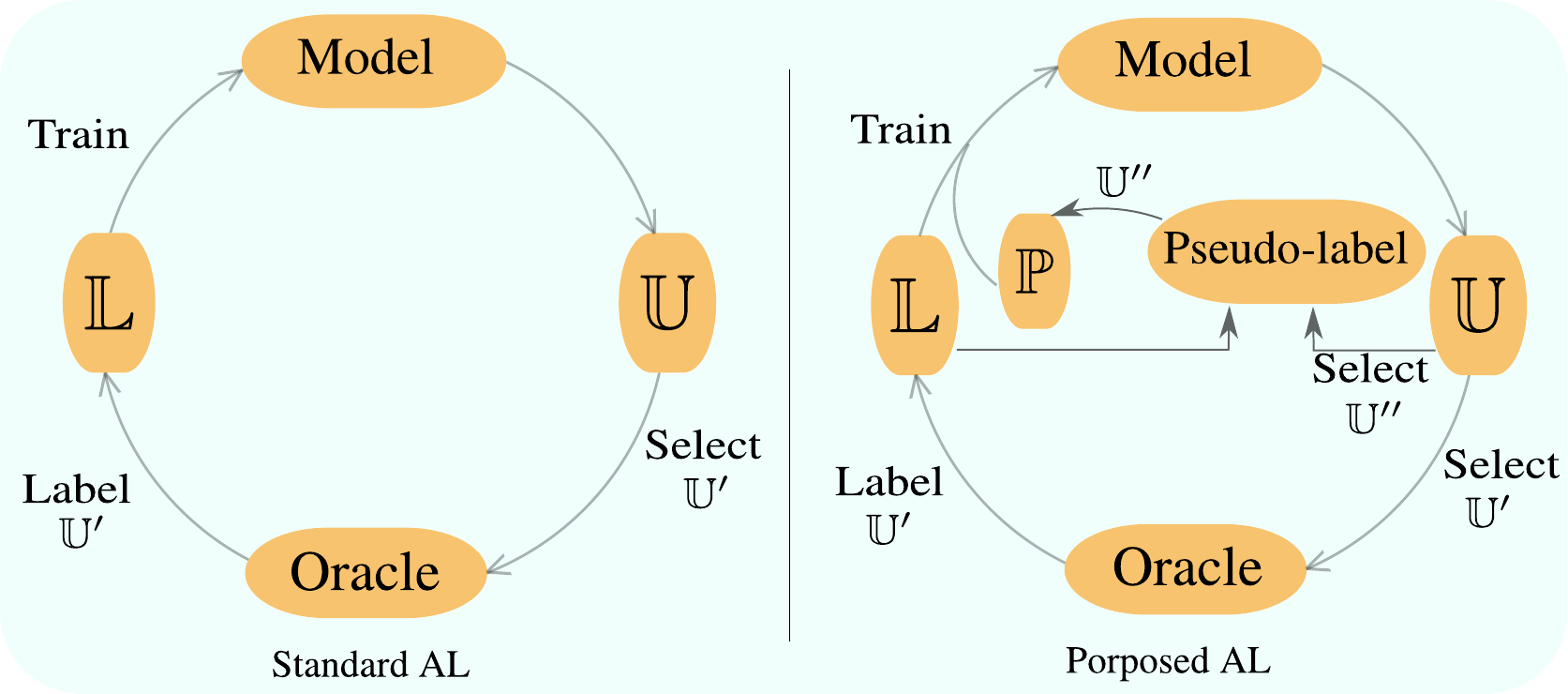}
  \caption{Proposed AL framework with weak annotator.}
  \label{fig:fig-proposal}
\end{figure}

Weakly-supervised learning (WSL) has recently emerged as a paradigm that relaxes the need for dense pixel-wise annotations \citep{rony2019weak-loc-histo-survey,zhou2017brief}. WSL techniques depend on the type of application scenario and annotation, such as global image-level labels  \citep{belharbi2019weakly,kim2017two,pathak2015constrained,teh2016attention,wei2017object}, scribbles \citep{Lin2016,ncloss:cvpr18}, points \citep{bearman2016}, bounding boxes \citep{dai2015boxsup,Khoreva2017} or global image statistics such as the target-region size \citep{bateson2019constrained,jia2017constrained,kervadec2019curriculum,kervadec2019constrained}. This paper focuses on learning using only image-level labels, which enables to classify an image while yielding pixel-wise scores (i.e., segmentations), thereby localizing the regions of interest linked to the image-class predictions.

Several CNN visualization and interpretation methods have recently been proposed, based on either perturbation, propagation or activation approaches, and allowing to localize the salient image regions responsible for a CNN’s predictions \citep{fu2020axiom}. In particular, WSL techniques \citep{rony2019weak-loc-histo-survey} rely on activation-based methods like CAM and, more recently, Gradient-weighted Class Activation Mapping (Grad-CAM), Grad-CAM++, Ablation-CAM and Axiom-based Grad-CAM \citep{fu2020axiom, lin2013network, PinheiroC15cvpr}. Trained with only global image-level annotations, these methods locate the regions of interest (ROIs) of the  corresponding class in a relatively inexpensive and accurate way. However, while these WSL techniques can provide satisfying results in natural images, they typically yield poor segmentations in relatively more challenging scenarios, for instance, histology data in medical image analysis \citep{rony2019weak-loc-histo-survey}.
We note two limitations associated with CAMs: (1) they are obtained in an unsupervised way (\ie without pixel-level supervision under an ill-posed learning problem \citep{choe2020evaluating}); and (2) they have low resolution. For instance, CAMs obtained from ResNet models \citep{heZRS16} have a resolution of ${1/32}$ of the input image. Interpolation is required to restore full image resolution. Both of these limitation with CAM-based methods lead to high false-positive rates, which may render them impractical \citep{rony2019weak-loc-histo-survey}.

Enhancing deep WSL models with pixel-wise annotation, as supported by a recent study in weakly-supervised object localization \citep{choe2020evaluating}, can improve localization and segmentation accuracy, which is the central goal of this paper. To do so, we introduce a deep WSL model that allows supervised learning for classification, and active learning for segmentation, with the latter providing more accurate and high-resolution masks. We assume that the images in the training set are globally annotated with image-class labels.  Relevant images are \emph{gradually} labeled at the pixel level through an oracle that respects a low  annotation-budget constraint. Consequently, this leads us to an active learning (AL) paradigm \citep{settles2009active}, where an oracle is requested to annotate pixels in a subset of samples.

Different sample-acquisition techniques have been successfully applied to deep AL for classification based on, e.g., certainty \citep{ducoffe2015qbdc,gal2017deep,kirsch2019batchbald} or representativeness \citep{kim2020task,sinha2019variational}. However, very few deep AL techniques were investigated in the context of segmentation \citep{gaur2016membrane,gorriz2017active,lubrano2019deep}. Most AL techniques focus mainly on the sampling criterion (Fig.\ref{fig:fig-proposal}, left) to populate the labeled pool using an oracle. During training, only the labeled pool is used, while the unlabeled pool is left dormant. Such an AL protocol may limit the accuracy of DL models under constrained oracle-supervision budget in real-world applications for multiple reasons:

\textbf{(1)} Standard AL protocols may be relevant to small/shallow models that can learn and provide reliable queries using a few training samples. Since training accurate DL models typically depends on large training sets, large numbers of queries may be needed to build reliable DL models, which may incur a high annotation cost.

\textbf{(2)} In most AL work, the experimental protocol starts with a large labeled pool, and acquires a large number of queries for sufficient supervision, neglecting the workload placed on the oracle.
This typically reaches a plateau-performance of a DL quickly, hampering a reliable study of the impact of different AL selection techniques. Moreover, model-based sampling techniques are inconsistent \citep{gaur2016membrane} in the sense that the model is used to query samples while it is still in an early learning stage.

\textbf{(3)} Segmentation and classification problems are associated with different properties and challenges, such as decision boundaries and uncertainty, which provide additional challenges to AL.  For instance, the class boundaries defined by different classification methods \citep{ducoffe2018adversarial,settles2009active,tong2001support} are not defined in the context of segmentation, making such a branch of methods inadequate for segmentation.

\textbf{(4)}
In critical fields such as medical imaging, acquiring a sample itself can be very expensive\footnote{
For instance, prior to a diagnosis of breast cancer from a histological sample, a patient undergoes a bilateral diagnostic mammogram and breast ultrasound that are interpreted by a radiologist, one to several needle biopsies (with low risks under ${1\%}$ of hematoma and wound infection) to further assess areas of concern, surgical consultation and pre-operative blood work, and surgical excision of the positive tissue for breast cancer cells. The biopsy and surgical tissues are processed (fixation, embedding in parraffin, H\&E staining) and interpreted by a pathologist. Depending on the cancer stage, the patient may undergo additional procedures. Therefore, accounting for all the steps required for breast cancer diagnosis from histological samples, a rough estimation of the final cost associated with obtaining a Whole Slide Image (WSI) is about \$400 (Canadian dollars, by 1999) \citep{will1999diagnostic}. Moreover, some cancer types are rare \citep{will1999diagnostic}, adding to the values of these samples. All these procedures are conducted by highly trained experts, with each procedure taking from a few minutes to an hour and requiring expensive specialized medical equipment.}.
The time and cost associated with each sample makes them valuable items. Such considerations may be overlooked for large-scale data sets with almost-free samples, as in the case of natural images. Given this high cost, keeping the unlabeled pool dormant during learning may be ineffective.

Based on the aforementioned arguments, we advocate that focusing solely on the sample acquisition and supervision pool may not be an efficient way to build
high-performing DL models in an AL framework for segmentation. Therefore, we consider augmenting the labeled pool using the model as a second source of annotation, in a self-learning fashion \citep{mao2020survey} (Fig.\ref{fig:fig-proposal}, right). This additional annotation might be less accurate (\ie, weak\footnote{Not to be confused with the weak annotation of data in weakly supervised learning frameworks.}) compared to the oracle that provides strong but expensive annotations. However, it is expected to fast-improve the model's performance \citep{mao2020survey}, while using a few oracle-annotated samples, reducing the annotation cost.

Our main contributions are the following.
\textbf{(1) Architecture design}: As an alternative to CAMs, we propose using a segmentation mask trained with pixel-level annotations, which yields more accurate and high-resolution ROIs. This is achieved through a convolutional architecture capable of simultaneously classifying and segmenting images, with the segmentation task trained using annotations acquired using an AL framework.
As illustrated in Fig.\ref{fig:fig-archi}, the architecture combines well-known DL models for classification (ResNet \citep{heZRS16}) and segmentation (U-Net \citep{Ronneberger-unet-2015}), although other architectures could also be used.
\textbf{(2) Active learning}: We augment the size of the labeled pool by weak-annotating a large number of unlabeled samples based on predictions of the DL model itself, providing a second source of annotation (Fig.\ref{fig:fig-proposal}). This enables rapid improvements of the segmentation accuracy, with less oracle-based annotation. Moreover, our method can be integrated on top of any sample-acquisition method.
\textbf{(3) Experimental study}: We conducted comprehensive experiments over two challenging benchmarks -- high-resolution medical images (histology GlaS data for colon cancer) and natural images (CUB-200-2011 for bird species). Our results indicate that, by simply using random sample selection, the proposed approach can significantly outperform state-of the-art CAMs and AL methods, with an identical oracle-supervision budget.

% ===================================================================================================
%
%                           RELATED WORK
%
% ===================================================================================================
\section{Related work}
\label{sec:related-work}

\noindent \textbf{Deep active learning:}
AL has been studied for a long time in machine learning, mainly for classification and regression, using linear models in particular  \citep{settles2009active}. Recently, there has been an effort to transfer such techniques to DL models for classification tasks by mimicking their intuition or by adapting them, taking into consideration model specificity and complexity. Such methods include, for instance, different mechanisms for uncertainty  \citep{beluch2018power,ducoffe2015qbdc,ducoffe2018adversarial,gal2017deep,kim2020task,kirsch2019batchbald,lakshminarayanan2017simple,wang2016cost,yoo2019learning} and representativeness estimation \citep{kim2020task,sener2018coreset,sinha2019variational}. However, most deep AL techniques are validated on synthetic, simple or tiny data, which does not explore their full potential in real applications.

While deep AL for classification is rapidly growing, deep AL models for segmentation are uncommon in the literature. In fact, the very few methods in the literature mostly focused on the direct application of deep AL classification methods. The limited research in this area may be explained by the fact that segmentation tasks bring challenges to AL, such as the additional spatial information and the fact that a segmentation mask lies in a much larger dimension than a classification prediction. In classification, AL often deals with one output that is used to drive queries \citep{huang2010active}. The spatial information in segmentation does not naturally provide a direct scoring function that can indicate the overall quality or certainty of the output. Most of deep AL methods for segmentation consider pixels as classification instances, and apply standard AL techniques to each pixel.

For instance, the authors of \citep{gaur2016membrane} exploit a variant of entropy-based acquisition at the pixel level, combined with a distribution-based term that encodes diversity using a complex hierarchical clustering algorithm over sliding windows, with application to microscopic membrane segmentation. Similarly, \citep{gorriz2017active,lubrano2019deep} apply Monte-Carlo dropout uncertainty \citep{gal2017deep} at the pixel level, with application to myelin segmentation using spinal cord and brain microscopic histology images. In \citep{roels2019cost}, the authors experiment with five acquisition functions of classification for a segmentation task, including entropy-based, core-set \citep{sener2018coreset}, k-mean and Bayesian \citep{gal2017deep} sampling,  with application to electron microscopy segmentation. Entropy-based methods seem to be dominant over multiple datasets. In \citep{yang2017suggestive}, the authors combine two sampling terms for histology image segmentation. The first employs bootstrapping over fully convolutional networks (FCN) to estimate uncertainty, where a set of FCNs are trained on different subsets of samples. The second term is a representation-based term that selects the most representative samples. This is achieved by solving an optimization of a generalization version of the maximum cover set problem \citep{feige1998threshold} using sample description extracted from an FCN. Despite the obtained promising results, this approach remains complex and impractical due to the use of bootstrapping over DL models and an optimization step. Moreover, the authors of \citep{yang2017suggestive} do not provide a comparison to other acquisition functions. The work in \citep{casanova2020} considers a specific case of AL using reinforcement learning for \emph{region-based} AL for segmentation, where only a selected region of the image is labeled. This method is adequate for data sets with large and unbalanced classes, such as street-view images. While the method in \citep{casanova2020} outperforms random and Bayesian \citep{gal2017deep} selection, surprisingly, it performs close to entropy-based selection.

\noindent \textbf{Weak annotators:}
The AL paradigm does not prohibit the use of unlabelled data for training \citep{settles2009active}, but it mainly constrains the oracle-labeling budget. The standard AL experimental protocol (Fig.\ref{fig:fig-proposal}, left) was inherited from AL of simple/linear ML models, and adopted in subsequent works. Budget-constrained oracle annotation may not be sufficient to build effective DL models, due to the lack of labeled samples. Furthermore, several studies in AL for classification have successfully leveraged the unlabelled data to provide additional supervision  \citep{lin2017active,long2008graph,vijayanarasimhan2012active,wang2016cost,zhou2010active,zhualsslharmo2003}.

For instance, the authors of \citep{lin2017active,wang2016cost} propose to pseudo-label a part of the unlabeled pool. The latter is selected using dynamic thresholding based on confidence, through the model itself, so as to learn a better embedding. Furthermore, a theoretical framework for AL using strong and weak annotators for classification task is introduced in \citep{zhang2015active}. Their results suggest that using multiple annotators can reduce the cost of oracle annotation, in comparison to one annotator. Multiple sources of annotations that include both strong and weak annotators were used in AL, crowd-sourcing, self-paced learning and other interactive learning scenarios for classification to help reducing the number of requests for the strong annotator \citep{kremer2018robust,malago2014online,mattsson2017active,murugesan2017active,urner2012learning,yan2016active,zhang2015active}.
Using the model itself for pseudo-annotation is motivated mainly by the success of deep self-supervised learning \citep{mao2020survey}.

\begin{wrapfigure}{R}{0.4\textwidth}
\centering
\includegraphics[scale=0.65]{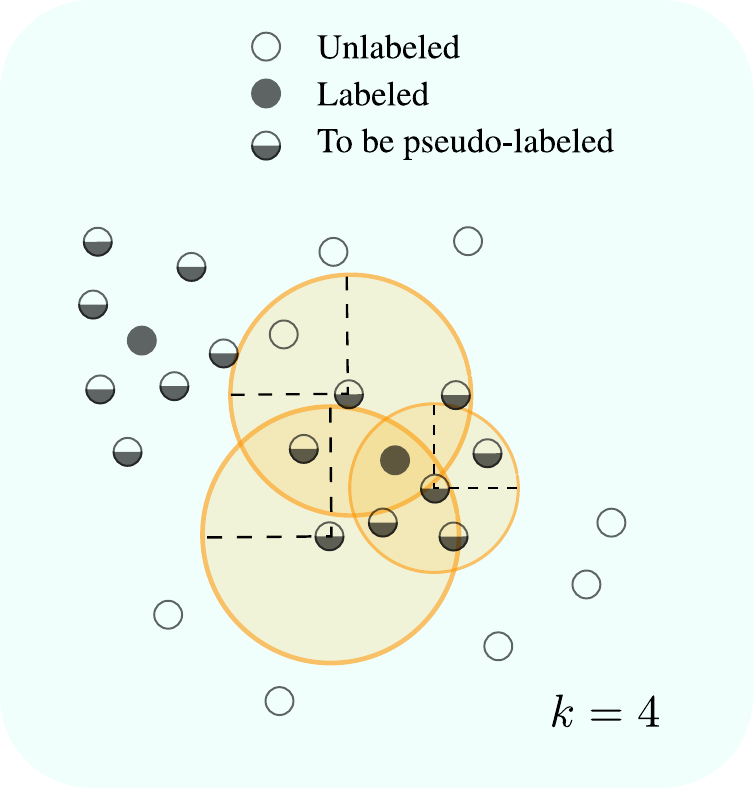}
\caption{The $k$-nn method for selecting $\mathbb{U}^{\prime\prime}$ subset to be speudo-labeled. Assumption to select $\mathbb{U}^{\prime\prime}$: since $\mathbb{U}^{\prime\prime}$ lives nearby supervised samples, it is more likely to be assigned good segmentation by the model.
We consider measuring $k$-nn for each \textbf{unlabeled} sample. In this example, using $k=4$ allows ${\abs{\mathbb{U}^{\prime\prime}} = 14}$. If $k$-nn is considered for each \textbf{labeled} sample: $\abs{\mathbb{U}^{\prime\prime}} = 8$. ${\abs{\cdot}}$ is the set cardinal. Note that ${k}$-nn is only considered between samples of the \emph{same class}.}
  \label{fig:fig-knn}
\vspace*{-0.1in}
\end{wrapfigure}

\noindent \textbf{Label Propagation (LP):}
Our approach is also related to LP methods \citep{bengiolabelprop2010,zhou2004learning,zhulp2002} for classification, which aim to label unlabeled samples using knowledge from the labeled ones (Fig.\ref{fig:fig-knn}). However, while LP propagates labels to unlabeled samples through an iterative process, our approach bypasses this using the model itself. In our case, the propagation is limited to the neighbors of labeled samples defined through $k$-nearest neighbors ($k$-nn) (Fig.\ref{fig:fig-knn}). Using $k$-nn has been also studied to combine AL and domain adaptation \citep{berlind2015active}, where the goal is to query samples from the target domain. Such an approach is connected to the recently developed core-set method for deep AL \citep{sener2018coreset}. Our method intersects with \citep{berlind2015active} only in the sense of predicting the labels to query samples using their labeled neighbors.

In contrast to state-of-the-art DL models for AL segmentation, we consider increasing the unlabeled pool through pseudo-annotated samples (Fig.\ref{fig:fig-proposal}, right). To this end, the model is used for pseudo-labeling samples within the neighborhood of samples with strong supervision (Fig.\ref{fig:fig-knn}). From a self-learning perspective, the works in \citep{lin2017active,wang2016cost} on face recognition are the closest to ours. While both rely on pseudo-labeling, they mainly differ in the sample selection for pseudo-annotation. In \citep{lin2017active,wang2016cost}, the authors considered model confidence, where samples with high confidence are pseudo-labeled, while low-confidence samples are queried. While this yields good results, it makes the overall method strongly dependent on model confidence. As we consider segmentation tasks, model-confidence is not well-defined. Moreover, using the expected pixel-wise values can be less representative for model confidence.

Our approach relies on the spatial assumption in Fig.\ref{fig:fig-knn}, where the samples to pseudo-label are selected to be near the labeled samples, and expected to have good pseudo-segmentations. This makes the oracle-querying technique independent from the pseudo-labeling method, giving more flexibility to the user. Our pseudo-labeled samples are added to the labeled pool, along with the samples annotated by the oracle. The underlying assumption is that, given a sample labeled by an oracle, the model is more likely to produce good segmentations of images located nearby that sample. Our assumption is verified empirically in the experimental section of this paper. This simple procedure enables to rapidly increase the number of pseudo-labeled samples, and helps improving segmentation performance under a limited amount of oracle-based supervision.

% ===================================================================================================
%                                      PROPOSAL
%
% ===================================================================================================
\section{Proposed approach}
\label{sec:proposal}

\begin{wrapfigure}{R}{0.4\textwidth}
\centering
 \includegraphics[scale=0.75]{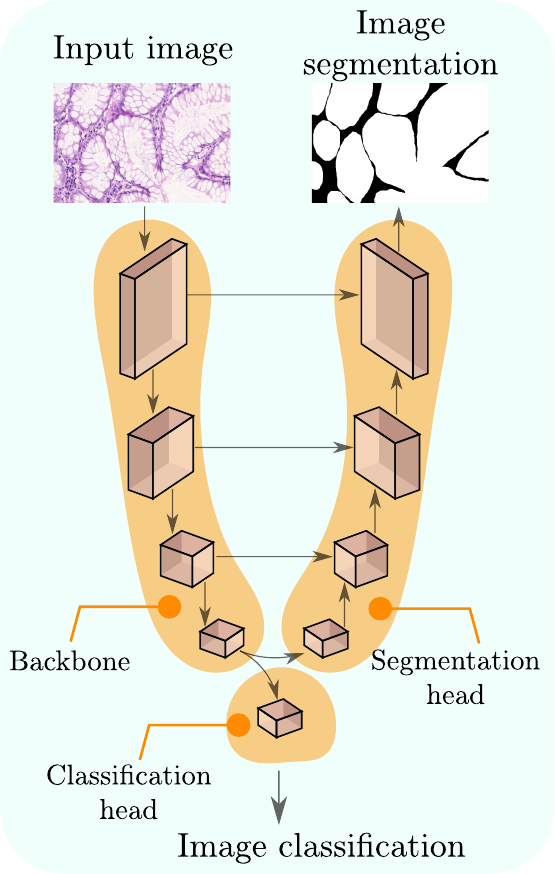}
 \caption{Our proposed DL architecture for classification and segmentation  composed of: (1) a shared \textbf{backbone} for feature extraction; (2) a \textbf{classification head} for the classification task; (3) and a \textbf{segmentation head} for the segmentation task with a U-Net style \citep{Ronneberger-unet-2015}. The latter merges representations from the backbone, while gradually upscaling the feature maps to reach the full image resolution for the predicted mask, similarly to the U-Net model.}
 \label{fig:fig-archi}
\vspace*{-0.1in}
\end{wrapfigure}

We consider an AL framework for training deep WSL models, where all the training images have class-level annotations, but no pixel-level annotations. Due to their high cost, pixel annotations are gradually acquired for training through oracle queries. It propagate pixel-wise knowledge encoded in the model though the labeled images.

Active learning training consists of sequential training rounds. At each round $r$, the total training set ${\mathbb{D}}$ that contains $n$ samples with unlabeled and labeled subsets (Fig.\ref{fig:fig-proposal}).
\textbf{(1) Unlabeled subset:} contains samples without pixel-wise annotation (unlabeled samples) ${\mathbb{U} = \{\bm{x}_i, y_i, -\-\}_{i=1}^u}$ where ${\bm{x} \in \mathcal{X}}$ is the input image, ${y}$ is its global label; and the pixel label is missing.
\textbf{(2) Labeled subset:} contains samples with full supervision ${\mathbb{L}=\{\bm{x}_i, y_i, \bm{m}_i\}_{i=1}^l}$ where ${\bm{m}}$ is the pixel-wise annotation of the sample. ${\mathbb{L}}$ is initially empty. It is gradually populated from ${\mathbb{U}}$ by querying the oracle using an acquisition function.
Let ${f(\cdot: \bm{\theta})}$ denotes a DL model that is able to classify and segment an image ${\bm{x}}$ (Fig.\ref{fig:fig-archi}). For clarity, and since we focus on the segmentation task, we omit the notation for the classification task (to simplify the presentation). Therefore, ${f(\bm{x})}$ refers to the predicted segmentation mask.
Let ${\mathbb{U}^{\prime} \subseteq \mathbb{U}}$ and ${\mathbb{U}^{\prime\prime} \subseteq \mathbb{U}}$ denote two subsets (Fig.\ref{fig:fig-proposal}), with ${ \mathbb{U}^{\prime} \cap \mathbb{U^{\prime\prime}} = \varnothing}$.
In our method, we introduce ${\mathbb{P}}$ as a subset holder for pseudo-labeled samples, which is initially empty and gradually replenished (Fig.\ref{fig:fig-proposal}, right).
To express the dependency of each subset on round ${r}$, we introduce the following notations: ${\mathbb{U}_r, \mathbb{L}_r, \mathbb{P}_r, \mathbb{U}^{\prime}_r, \mathbb{U}^{\prime\prime}_r}$.  The samples in ${\mathbb{P}_r}$ are denoted as ${\{\bm{x}_i, y_i, \hat{\bm{m}}_i\}}$. The following holds: ${\forall r: \mathbb{D} = \mathbb{L}_r \cup \mathbb{U}_r \cup \mathbb{P}_r}$.

Alg.\ref{alg:alg-0} describes the overall AL process with our pseudo-annotation method.
First, ${\mathbb{U}^{\prime}_r}$ is queried, then labeled by the oracle, and added to ${\mathbb{L}_r}$.
Using $k$-nn, ${\mathbb{U}^{\prime\prime}_r}$ is selected based on their proximity to ${\mathbb{L}_r}$ (Fig.\ref{fig:fig-knn}); and pseudo-labeled by the model, then added to ${\mathbb{P}_r}$. To fast-increase the size of ${\mathbb{L}_r}$, ${\mathbb{P}_r}$ is protected from being queried for the oracle until it is inevitable. In the \emph{latter case}, queried samples from ${\mathbb{P}_r}$ are used to fill ${\mathbb{U}^{\prime}}$; and they are no longer considered pseudo-labeled since they will be assigned the oracle annotation.

To measure image similarity for the $k$-nn method, we used the color distribution to describe image content. This can be a flexible descriptor for highly unstructured images such as histology images. Note that the ${k}$-nn method is considered \emph{only} for pairs of samples of the \emph{same class}. The underlying assumption is that samples of the same class, with similar color distributions, are likely to contain relatively similar objects. Consequently, labeling representative samples could be a proxy for supervised learning based on the underlying data distribution. This can increase the likelihood of the model to provide relatively good segmentations of the other samples. The proximity between two images ${(\bm{x}_i, \bm{x}_j)}$ is measured using the Jensen-Shannon divergence between their respective color distributions (measured as normalized histograms). For an image with multiple color planes, the similarity is formulated as the sum of similarities, one for each plane.

At round $r$, the queried and pseudo-labeled samples are both used in training by optimizing the following loss function:
\begin{equation}
\label{eq:eq-1}
    \min_{\bm{\theta}} \sum_{\bm{x}_i \in \mathbb{L}_{r-1}} \mathcal{L}(f(\bm{x}_i), \bm{m}_i) + \lambda \sum_{\bm{x}_i \in \mathbb{P}_{r-1}} \mathcal{L}(f(\bm{x}_i), \hat{\bm{m}}_i),
\end{equation}
where ${\mathcal{L}(\cdot, \cdot)}$ is a segmentation loss, and ${\lambda}$ a positive scalar. Eq.(\ref{eq:eq-1}) corresponds to training the model (Fig.\ref{fig:fig-archi}) solely for the segmentation task. Simultaneous training for classification and segmentation in this AL setup is avoided due to the unbalance between the number of samples that are labeled globally and at the pixel level. Therefore, we consider training the model for classification first. Then, we freeze the classifier parameters. Training for the segmentation tasks is resumed later. This yields the best classification performance, and allows a better study of the impact of the queried samples on the segmentation task.

Considering the relation of our method and label propagation algorithm \citep{bengiolabelprop2010,zhou2004learning,zhulp2002}, we refer to our proposal as Label\_prop.

\begin{center}
\begin{minipage}{0.6\linewidth}
\IncMargin{0.04in}
% \removelatexerror% Nullify \@latex@error
\begin{algorithm}[H]
    \SetKwInOut{Input}{Input}
    %\small
    \Input{
    ${\mathbb{P}_0 = \mathbb{L}_0 = \varnothing}$
    \\
    ${\bm{\theta}^0}$: Initial parameters of ${f}$ trained on the classification task.
    \\
    ${\texttt{maxr}: \textrm{Maximum number of AL rounds}}$.
    }
    \vspace{0.1in}
    Select ${\mathbb{U}^{\prime}_0}$ randomly and label them by an oracle. \\
    \vspace{0.025in}
    $  \mathbb{L}_0 \la \mathbb{U}^{\prime}_0$. \\
    \For{${r \in 1 \cdots \texttt{maxr}}$}  %\texttt{\%Iteratively replenish ${\mathbb{P}_k}$}
    {
        %$\minibatch{\ell} \la \textrm{sample mini-batch}$\;
        ${\bm{\theta} \la \bm{\theta}^0}$. \\
        \vspace{0.03in}
        Train $f$ using ${\mathbb{L}_{r-1}}$ \colorbox{mybluelight}{${\cup\; \mathbb{P}_{r-1}}$} and the loss in Eq. (\ref{eq:eq-1}). \\
        \vspace{0.03in}
        Select ${\mathbb{U}^{\prime}_r}$ and label them by an oracle. \\
        \vspace{0.03in}
        $  \mathbb{L}_r \la \mathbb{L}_{r-1} \cup \mathbb{U}^{\prime}_r$. \\
        \vspace{0.03in}
        \colorbox{mybluelight}{Select ${\mathbb{U}^{\prime\prime}_r}$.} \\
        \vspace{0.03in}
        \colorbox{mybluelight}{$ \mathbb{P}_r  \la  \mathbb{P}_{r-1} \cup \mathbb{U}^{\prime\prime}_r$.} \\  % or \;
        \vspace{0.03in}
        \colorbox{mybluelight}{Pseudo-label ${\mathbb{P}_r}$.}
    }
    \vspace{0.1in}
    \caption{Standard AL procedure and our proposal. The extra instructions associated with our method are indicated with a \colorbox{mybluelight}{blue background}.
    }
    \label{alg:alg-0}
\end{algorithm}
\DecMargin{0.04in}
\end{minipage}
\end{center}

% ===================================================================================================
%
%                                      EXPERIMENTS
%
% ===================================================================================================
\section{Results and discussion}
\label{sec:experiments}

%This section describes the experimental protocol and the obtained results.

 \subsection{Experimental methodology:}

 \begin{figure}[!b]
  \centering
  \includegraphics[width=0.99\linewidth]{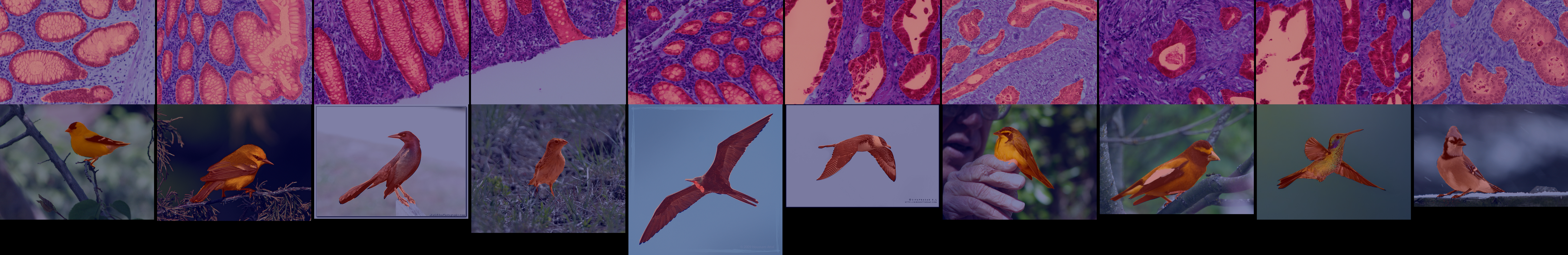}
    \caption{\textbf{Top row}: GlaS dataset \citep{sirinukunwattana2017gland}. \textbf{Bottom row}: CUB dataset \citep{WahCUB2002011}. (Best visualized in color.)}
  \label{fig:fig-datasets}
\end{figure}

% ===================================================================================================
%                                           DATASETS
% ===================================================================================================
\noindent \textbf{a) Datasets.}
For evaluation, datasets should have global and pixel-wise annotation. We consider two public datasets including both medical (histology) and natural images (Fig.\ref{fig:fig-datasets}).
\begin{inparaenum}[(1)]
 % glas
  \item \textbf{GlaS dataset}: This dataset contains histology images for colon cancer diagnosis\footnote{GlaS: \href{https://warwick.ac.uk/fac/sci/dcs/research/tia/glascontest}{warwick.ac.uk/fac/sci/dcs/research/tia/glascontest}.} \citep{sirinukunwattana2017gland}. It includes 165 images derived from 16 Hematoxylin and Eosin (H\&E) histology sections of two grades (classes): benign and malignant. It is divided into 84 samples for training and 80 samples for testing.
  The ROIs to be segmented are the glandes.
  % future
%   \item \todoit{\textbf{batch-based CAMELYON16}:.}
  % cub
  \item \textbf{CUB-200-2011 dataset (CUB)}\footnote{CUB: \href{http://www.vision.caltech.edu/visipedia/CUB-200-2011.html}{www.vision.caltech.edu/visipedia/CUB-200-2011.html}} \citep{WahCUB2002011} is a dataset for bird species with ${11,788}$ samples ($5,994$ for training and $5,794$ for testing) and ${200}$ species. The ROIs to be segmented are the birds.
  % oxf
  \end{inparaenum}
  In GlaS and CUB datasets, we randomly select $80\%$ of the training samples for effective training, and $20\%$ for validation (with full supervision) to perform early stopping. The splits are identical to the ones used in \citep{belharbi2019unimoconstraints,rony2019weak-loc-histo-survey} (split 0, fold 0), and are publicly available.

% ===================================================================================================
%                                       ACTIVE LEARNING SETUP
% ===================================================================================================
%\subsection{Active learning setup}
\noindent \textbf{b) Active learning setup.}
To assess the performance of different AL acquisition methods, we consider a realistic scenario with respect to the number of samples to be labeled at each AL round, accounting for the load imposed on the oracle.
Therefore, only a few samples are selected at each round for oracle annotation, and  ${\mathbb{L}}$ is slowly replenished. This allows better comparison between AL selection techniques since we spend more time
in a phase where ${\mathbb{L}}$ holds a few samples. Such a phase allows to better measure the impact of the selected samples. Filling  ${\mathbb{L}}$ quickly brings the model's performance to a plateau that hides the impact of newly selected samples.
The initial replenishment ($r=1$) is achieved by randomly selecting a few samples. The same samples are used for all AL techniques at round $r=1$ for a fair comparison. To avoid any bias from selecting unbalanced classes that can directly affect the segmentation performance, and hinder AL evaluation, the same number of samples is selected from each class (since the global annotation is known beforehand for all the samples). Note that the oracle is used only to provide pixel-wise annotation. Tab.\ref{tab:tab0} provides the selection details.
\begin{table}[t!]
\renewcommand{\arraystretch}{1.3}
\caption{Number of samples selected for the oracle per round.}
\label{tab:tab0}
\centering
   \resizebox{0.7\linewidth}{!}{
\begin{tabular}{l|ccc}
    Dataset  &  \makecell[c]{\#selected samples \\ \emph{per-class} ($r=1$)} & \makecell[c]{\#selected samples \\ \emph{per-class} ($r > 1$)} & \makecell[c]{Max AL rounds \\ (\texttt{maxr} in Alg.\ref{alg:alg-0})}\\
    \toprule
    GlaS   &   $4$   & $1$  & $25$ \\
    CUB   &   $1$  & $1$    & $20$ \\
    \bottomrule
\end{tabular}
}
\end{table}
%

% ===================================================================================================
%                                           EVALUATION
% ===================================================================================================
\noindent \textbf{c) Evaluation.}
%\subsection{Evaluation}
We report the classification accuracy obtained by the classification head (Fig.\ref{fig:fig-archi}). Average Dice index is used to measure the segmentation quality at each AL round forming a Dice index curve over all the rounds. To better assess the \emph{dominance} of each method \citep{settles2009active}, the Area Under the Dice index Curve is used (AUC). This provides a fair indicator of the dominant curve, but contrasts with standard AL works, where one or multiple specific operating points in the curve are selected (leading to a biased and less accurate protocol). The average and standard deviation of Dice index curve and AUC metric are reported based on 5 replications of a complete AL session, using a different seed for each session. An AL session across different methods uses the same seed. %{\tiny ${\blacksquare}$}

While our approach, referred to as (Label\_prop), can operate on top of any AL selection criterion, we demonstrate its efficiency using simple random selection, which is often a baseline for AL experiments. Note that our pseudo-annotations are obtained from the segmentation head shown in Fig.\ref{fig:fig-archi}. Our method is compared to three different AL selection approaches for segmentation:
\textbf{(1) random selection (Random)}: the samples are randomly selected;
\textbf{(2) entropy-based selection (Entropy)}: the scoring function per sample is the average entropy at the pixel level \citep{gaur2016membrane}. Samples with high entropy are selected; and
\textbf{(3) Monte-Carlo dropout uncertainty (MC\_Dropout)}: we use Monte-Carlo dropout \citep{gorriz2017active,lubrano2019deep} at the pixel level to compute the uncertainty score per sample. Samples are forwarded ${50}$ times in the model, where dropout is set to ${0.2}$ \citep{gorriz2017active,lubrano2019deep}. Then, the pixel-wise variance is estimated. Samples with high mean variance are selected.
%{\tiny ${\blacksquare}$}
%

\noindent  \textbf{Lower bound performance (WSL)}: We consider the segmentation performance obtained by WSL method as a lower bound. It is trained using only global annotation. CAMs are used to extract the segmentation mask. WILDCAT method is considered \citep{durand2017wildcat} (Fig.\ref{fig:fig-archi}) at the classification head to obtain the CAMs. For WSL method, a pre-trained model over ImageNet \citep{imagenetcvpr09} is used to initialize the weights of the backbone, which is then fined-tuned.
The model is trained over the entire dataset, where samples are labeled globally only.
The obtained classifier using seed=0 is frozen and used as a backbone for \emph{all} the other methods.

\noindent \textbf{Upper bound performance (Full\_sup)}: Fully supervised segmentation is considered as an upper bound on performance. The model in Fig.\ref{fig:fig-archi} is trained for segmentation only using the entire dataset, where samples are labeled at the pixel level.
%{\tiny ${\blacksquare}$}

For a fair comparison, all the methods are trained using the same hyper-parameters over the same dataset. WSL and Full\_sup methods have minor differences. Due to space limitation, all the hyper-parameters are presented in the supplementary material.
In Alg.\ref{alg:alg-0}, notice that for our method, ${\mathbb{P}_r}$ is not used at the current round $r$ but until the next round ${r+1}$. To take advantage of ${\mathbb{P}_r}$ at round $r$, instructions from line-4 to line-10 are repeated twice in the provided results.

% ===================================================================================================
%                                           RESULTS
% ===================================================================================================
\subsection{Results}
\label{sub:results}

\begin{figure}[htbp]
  \centering
 \begin{subfigure}[b]{.5\linewidth}
  \centering
  \includegraphics[scale=0.39]{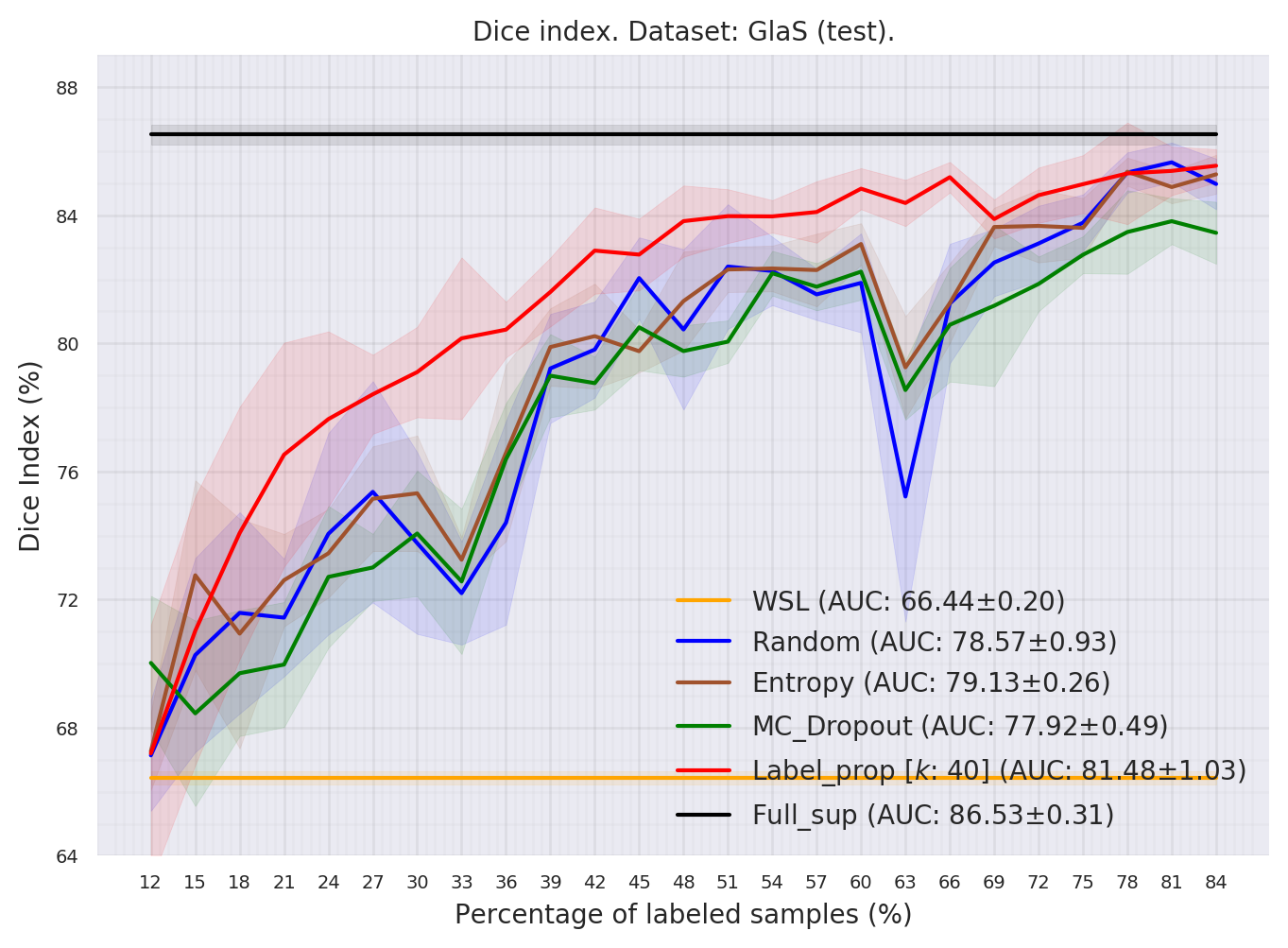}
  \caption{}
  \label{fig:fig-al-0}
 \end{subfigure}%
 \begin{subfigure}[b]{.5\linewidth}
  \centering
  \includegraphics[scale=0.39]{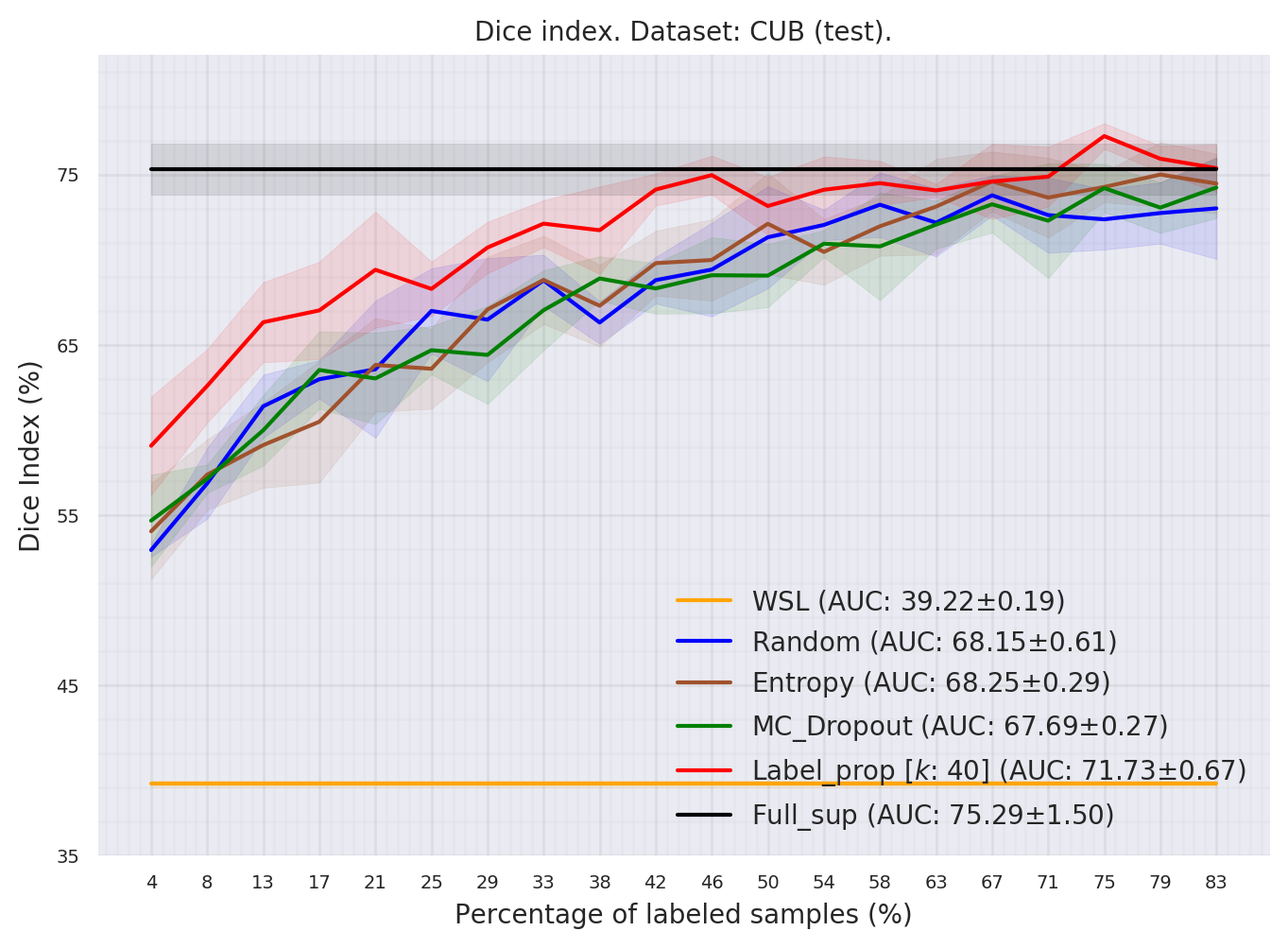}
  \caption{}
  \label{fig:fig-al-1}
 \end{subfigure}%
\caption{Average Dice index of the proposed and baseline methods over test sets.
(\protect\subref{fig:fig-al-0}) GlaS.
(\protect\subref{fig:fig-al-1}) CUB.
}
\label{fig:fig-al-results}
\end{figure}

\begin{table}[t!]
\renewcommand{\arraystretch}{1.3}
\caption{Classification accuracy over of the proposed deep WSL model on GlaS and CUB test datasets.}
\label{tab:tab-cl-acc}
\centering
   \resizebox{.5\linewidth}{!}{
\begin{tabular}{l|ccc}
    Dataset  &  \multicolumn{1}{c}{GlaS} & \multicolumn{1}{c}{CUB}\\
    \toprule
    \makecell{Classification \\ accuracy (\%)}   &   $99.50 \pm 0.61$
                                                 &   $73.22 \pm 0.19$
                                                 \\
    \bottomrule
\end{tabular}
}
\end{table}

\begin{table}[h!]
\renewcommand{\arraystretch}{1.3}
\caption{Average AUC and standard deviation (Fig.\ref{fig:fig-al-results}) for Dice index performance over GlaS and CUB test sets.}
\label{tab:tab-seg-perf}
\centering
   \resizebox{.6\linewidth}{!}{
\begin{tabular}{l|ccc}
    Dataset  &  \multicolumn{1}{c}{GlaS} & \multicolumn{1}{c}{CUB} \\
    \toprule
    \toprule
    WSL   &    $66.44 \pm 0.20$  &   $39.22 \pm 0.19$  \\
    \midrule
    Random   &    $78.57 \pm 0.93$  &   $68.15 \pm 0.61$  \\
    Entropy   &   $79.13 \pm 0.26$  &   $68.25 \pm 0.29$  \\
    MC\_Dropout   &   $77.92 \pm 0.49$  &   $67.69 \pm 0.27$  \\
    \rowcolor{mybluelight}
    \textbf{Label\_prop (ours)}   &   $\bm{81.48 \pm 1.03}$  &   $\bm{71.73 \pm 0.67}$ \\
    \midrule
    Full\_sup   &   $86.53 \pm 0.31$  &   $75.29 \pm 1.50$  \\
    \bottomrule
\end{tabular}
}
\end{table}

\begin{table*}[h!]
\renewcommand{\arraystretch}{1.3}
\caption{Readings of Dice index (mean $\pm$ standard deviation) from Fig.\ref{fig:fig-al-results} over test set for the \textbf{first 5 queries} formed by each method. We start from the second query since the first query is random but identical for all methods.}
\label{tab:tab-dice-q}
\centering
   \resizebox{0.9\linewidth}{!}{
\begin{tabular}{l|cccccc}
   \textbf{Queries}  &  \textbf{q2} & \textbf{q3} & \textbf{q4} & \textbf{q5} & \textbf{q6} \\
   \toprule
   \toprule
   \multicolumn{6}{c}{GlaS} \\
    \toprule
    WSL   &    \multicolumn{4}{c}{$66.44 \pm 0.20$}  \\
    \midrule
    Random   &   $70.26 \pm 3.02$ &   $71.58 \pm 3.14$ & $71.43 \pm 1.83$ & $74.05 \pm 3.14$ &    $75.36 \pm 3.45$\\
    Entropy   &   $\bm{72.75 \pm 2.96}$ &   $70.93 \pm 3.58$ & $72.60 \pm 1.44$ & $73.44 \pm 1.38$ &   $75.15 \pm 1.63$\\
    MC\_Dropout   &   $68.44 \pm 2.89$  &   $69.70 \pm 1.96$ &   $69.97 \pm 1.95$ & $72.71 \pm 2.21$ & $73.00 \pm 1.04$\\
    \rowcolor{mybluelight}
    \textbf{Label\_prop (ours)}   &   $71.02 \pm 4.19$  &   $\bm{74.07 \pm 3.93}$ &   $ \bm{76.52 \pm 3.49}$ & $\bm{77.63 \pm 2.73}$ & $\bm{78.41 \pm 1.23}$\\
   % \midrule
    Full\_sup   &   \multicolumn{4}{c}{$86.53 \pm 0.31$}\\
    \toprule  \toprule
   \multicolumn{6}{c}{CUB} \\
    \toprule
    WSL   &    \multicolumn{4}{c}{$39.22 \pm 0.18$}  \\
    \midrule
    Random   &   $56.86 \pm 2.07$ &   $61.39 \pm 1.85$ & $62.97 \pm 1.13$ & $63.56 \pm 4.02$ &    $66.56 \pm 2.50$\\
    Entropy   &   $53.37 \pm 2.06$ &   $59.11 \pm 2.50$ & $60.48 \pm 3.56$ & $63.81 \pm 2.75$ &   $63.59 \pm 2.34$\\
    MC\_Dropout   &   $57.13 \pm 0.83$  &   $59.98 \pm 2.06$ &   $63.52 \pm 2.26$ & $63.02 \pm 2.68$ & $64.68 \pm 1.41$\\
    \rowcolor{mybluelight}
    \textbf{Label\_prop (ours)}  &   $\bm{62.58 \pm 2.15}$  &   $\bm{66.32 \pm 2.34}$ &   $ \bm{67.01 \pm 2.85}$ & $\bm{69.40 \pm 3.40}$ & $\bm{68.28 \pm 1.60}$\\
    \midrule
    Full\_sup   &   \multicolumn{4}{c}{$75.29 \pm 1.50$}\\
    \bottomrule
    \bottomrule
\end{tabular}
}
\end{table*}

We report the classification and segmentation performances following the training the proposed deep WSL model in Fig.\ref{fig:fig-archi}. Tab.\ref{tab:tab-cl-acc} reports the Classification accuracy of the classification head using WSL, which is
close to the results reported in \citep{belharbi2019weakly,rony2019weak-loc-histo-survey}. The results of GlaS suggest that it is an easy dataset for classification.

The segmentation results are reported in Tabs. \ref{tab:tab-dice-q} and \ref{tab:tab-seg-perf}, and in Fig \ref{fig:fig-al-results}.

Fig. \ref{fig:fig-al-0} compares Dice accuracy on the \textbf{GlaS dataset}. On the latter, we observe that adding more labels increases Dice index for all AL methods, yielding, as expected, better performance than the WSL method.  Reading from Tab.\ref{tab:tab-dice-q}, randomly labeling only 4 samples per class enables to easily outperform WSL. This means that using our approach in Fig.\ref{fig:fig-archi}, with limited supervision, can lead to more accurate masks compared to using CAMs in the WSL method. From Fig.\ref{fig:fig-al-0}, one can also observe that Random, Entropy, and MC\_Dropout methods grow relatively in the same way, leading to the same overall performance, with the Entropy method slightly ahead. Considering the overall behavior of the curves, one may conclude that using advanced selection techniques such as MC\_Dropout and Entropy provides an accuracy similar to simple random selection. On the one hand, since both methods have shown substantial improvements in AL for classification, and based on the results in Fig.\ref{fig:fig-al-0}, one may conclude that all samples are equivalently informative for the model. Therefore, there is no better order to acquire them. On the other hand, using simply random selection and pseudo-labeled samples allowed our method to substantially improve the overall performance,  demonstrating the benefits of self-learning.

Fig.\ref{fig:fig-al-1} and Tab.\ref{tab:tab-dice-q} compare Dice accuracy on the \textbf{CUB dataset}, where labeling only one sample per class yielded a large improvement in Dice index, in comparison to WSL. Adding more samples increases the performance of all the methods. One can observe similar pattern as for GlaS: Random, Entropy and MC\_Dropout methods yield similar curves, while the AUC performances of Random and Entropy methods are similar, and slightly ahead of MC\_Dropout.
Similar to GlaS analysis, and based on the results of these three methods, one can conclude that none of the methods for ordering the samples is better than simple random selection. Using self-labeled samples in our method shows again its benefits. Simple random selection combined with self-annotation yields an overall best performance. Using two datasets, our empirical results suggest that self-learning, under limited oracle-annotation, has the potential to provide a reliable second source of annotation, which can efficiently enhance model performance, while using simple sample acquisition techniques.

\begin{figure}[h!]
  \centering
  \includegraphics[width=.7\linewidth]{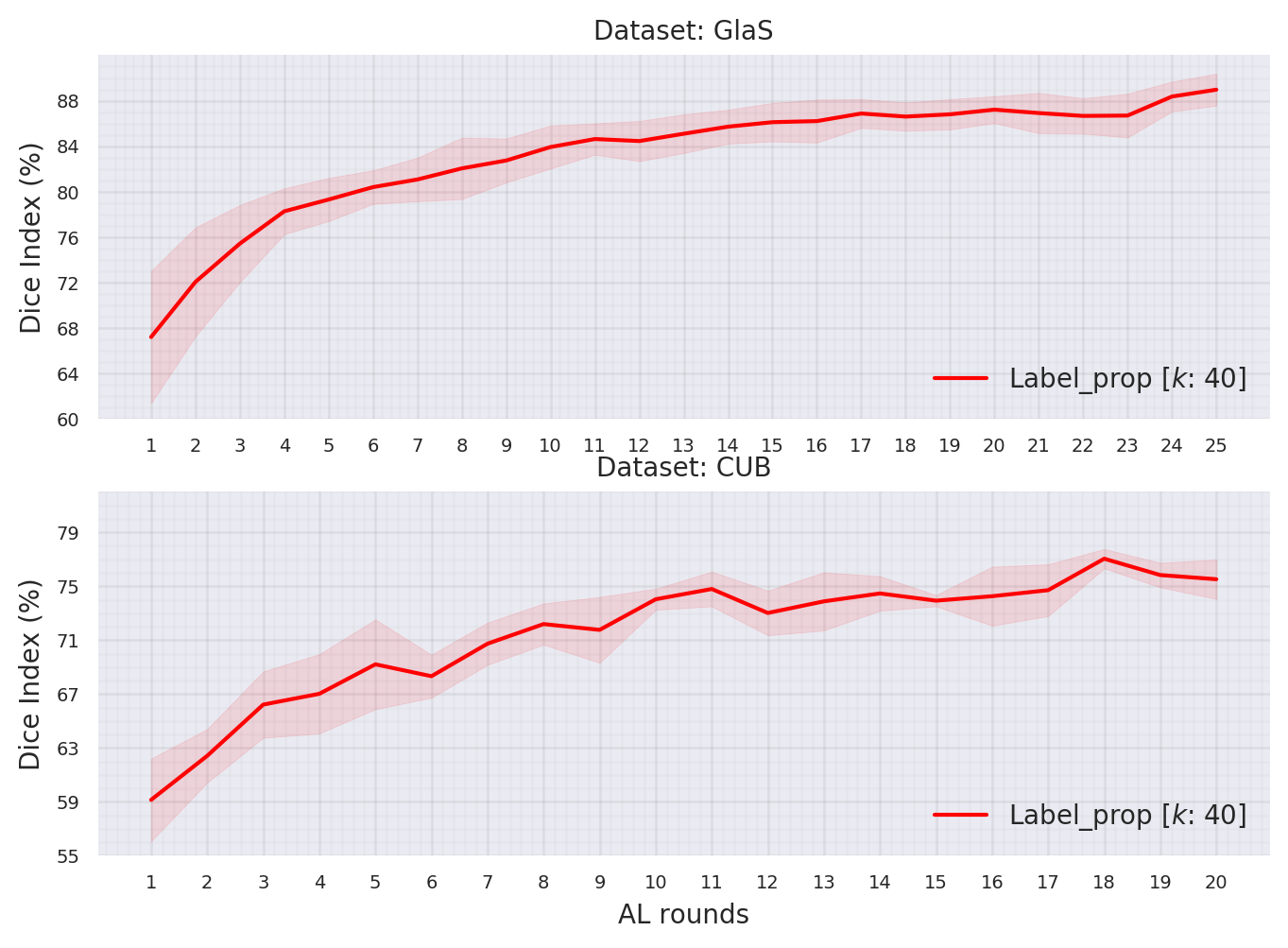}
  \caption{Average Dice index over the pseudo-labeled samples of our method in \textbf{each} AL round.}
  \label{fig:fig-dice-pseudo-labeled}
\end{figure}

\noindent \textbf{Pseudo-annotation performance}. Furthermore, the proposed approach is assessed on the pseudo-labeled samples at each AL round. Fig.\ref{fig:fig-dice-pseudo-labeled} shows that the model provides good segmentations at the initial rounds. Then, the more supervision, the more accurate the pseudo-segmentation, as expected. This figure shows the interest and potential of self-learning in segmentation, and confirms our assumption that samples near the labeled ones are likely to achieve accurate pseudo-segmentation by the model.

\noindent \textbf{Hyper-parameters}. Our approach requires two main hyper-parameters: ${ k \ \text{and } \lambda}$. We conducted an ablation study over ${k}$ on GlaS dataset, and over ${\lambda}$ on both datasets. Results, which are presented in the supplementary material, suggest that our method is less sensitive to ${k}$. ${\lambda}$ plays an important role, and
based on our study, we recommend using small values of this weighting parameter. In our experiments, we used $\lambda=0.1$ for Glas and $\lambda = 0.001$ for CUB. We set ${k=40}$. We note that hyper-parameter tuning in AL is challenging due to the change of the size of the data set, which in turn changes the training dynamics. In all the experiments, we used fixed hyper-parameters across the AL rounds.
Fig.\ref{fig:fig-dice-pseudo-labeled} suggests that a dynamic ${\lambda(r)}$ that is increased through AL rounds could be more beneficial. However, this requires a principled update protocol for ${\lambda}$, which was not explored in this work. Nonetheless, using a fixed value seems to yield promising results overall.

% removed due to space limitation.
\noindent \textbf{Supplementary material}. Due to space limitation, we deferred the hyper-parameters used in the experiments, results of the ablation study, visual results for the similarity measure and examples of predicted masks to the supplementary materials.

% ===================================================================================================
%
%
%                                      CONCLUSION
%
%
% ===================================================================================================

\section{Conclusion}
\label{sec:conclusion}

Deep WSL models trained with global image-level annotations can play an important role in CNN visualization and interpretability. However, they are prone to high false-positive rates, especially for challenging images, leading to poor segmentations.
To alleviate this issue, we considered  using pixel-wise supervision provided gradually through an AL framework. This annotation is integrated into training using an adequate deep convolutional model that allows supervised learning for both tasks: classification and segmentation. Through a few pixel-supervised samples, such a design is intended to provide full-resolution and more accurate masks compared to standard CAMs, which are trained without pixel supervision and often provide coarse resolution. Therefore, it enables a better CNN visualization and interpretation of CNN predictions.
Furthermore, and unlike standard deep AL methods that focus solely on the acquisition function, we considered using self-learning as a second source of supervision to fast-improve the model segmentation.
Evaluating our method using a realistic AL protocol over two challenging benchmarks, our results indicate that:
(1) using a \emph{few} supervised samples, the proposed architecture yielded more accurate segmentations compared to CAMs, with a large margin using different AL methods. Thus, it provides a solution to enhance pixel-wise predictions
in real-world visual recognition applications.
(2) using self-learning with random selection yielded substantial improvements. Self-learning under a limited oracle-budget can, therefore, provide a cost-effective alternative to standard AL protocols, where most of the effort is spent on the acquisition function.

\section*{Acknowledgment}
This research was supported in part by the Canadian Institutes of Health Research, the Natural Sciences and Engineering Research Council of Canada, Compute Canada, MITACS, and the Ericsson Global AI Accelerator Montreal.

\clearpage
\newpage

\appendices

% ===================================================================================================
%
%
%                                          SUPP-MATERIAL
%
%
% ===================================================================================================

\section{Supplementary material for the experiments}
Due to space limitation, we provide in this supplementary material detailed hyper-parameters used in the experiments, results of the ablation study, visual results to the similarity measure, and examples of predicted masks.

\subsection{Training hyper-parameters}
\label{subsec:hyper-params}
Tab.\ref{tab:tabx-tr-hyper-params} shows the used hyper-parameters in all our experiments.

\begin{table}[h!]
\renewcommand{\arraystretch}{1.3}
\caption{Training hyper-parameters.}
\label{tab:tabx-tr-hyper-params}
\centering
\resizebox{0.7\linewidth}{!}{
\begin{tabular}{lccc}
    Hyper-parameter  &  GlaS  & CUB\\
    \toprule
    Model backbone  & \multicolumn{2}{c}{ResNet-18 \citep{heZRS16}}\\
    WILDCAT \citep{durand2017wildcat}: && \\
    ${\alpha}$ &  \multicolumn{2}{c}{${0.6}$} \\
    ${kmin}$ &  \multicolumn{2}{c}{${0.1}$} \\
    ${kmax}$ &  \multicolumn{2}{c}{${0.1}$} \\
    modalities &  \multicolumn{2}{c}{${5}$} \\
    \midrule
    Optimizer &  \multicolumn{2}{c}{SGD}\\
    Nesterov acceleration & \multicolumn{2}{c}{True}\\
    Momentum & \multicolumn{2}{c}{$0.9$} \\
    Weight decay & \multicolumn{2}{c}{$0.0001$}\\
    Learning rate (LR) & ${0.1}$ (WSL: $10^{-4}$) & ${0.1}$ (WSL: $10^{-2}$)\\
    LR decay & ${0.9}$  & ${0.95}$ (WSL: ${0.9}$) \\
    LR frequency decay & $100$ epochs & $10$ epochs  \\
    Mini-batch size & ${20}$ & ${8}$ \\
    Learning epochs & ${1000}$ & ${30}$ (WSL: ${90}$) \\
    \midrule
    Horizontal random flip & \multicolumn{2}{c}{True} \\
    Vertical random flip & True & False  \\
    % Random color jittering & ? & ? \\
    Crop size & \multicolumn{2}{c}{${416 \times 416}$}\\
    \midrule
    ${k}$ & \multicolumn{2}{c}{${40}$}\\
    ${\lambda}$ & ${0.1}$ & ${0.001}$ \\
    \bottomrule
\end{tabular}
}
\end{table}

\begin{figure}[h!]
\centering
  \centering
  \includegraphics[scale=0.5]{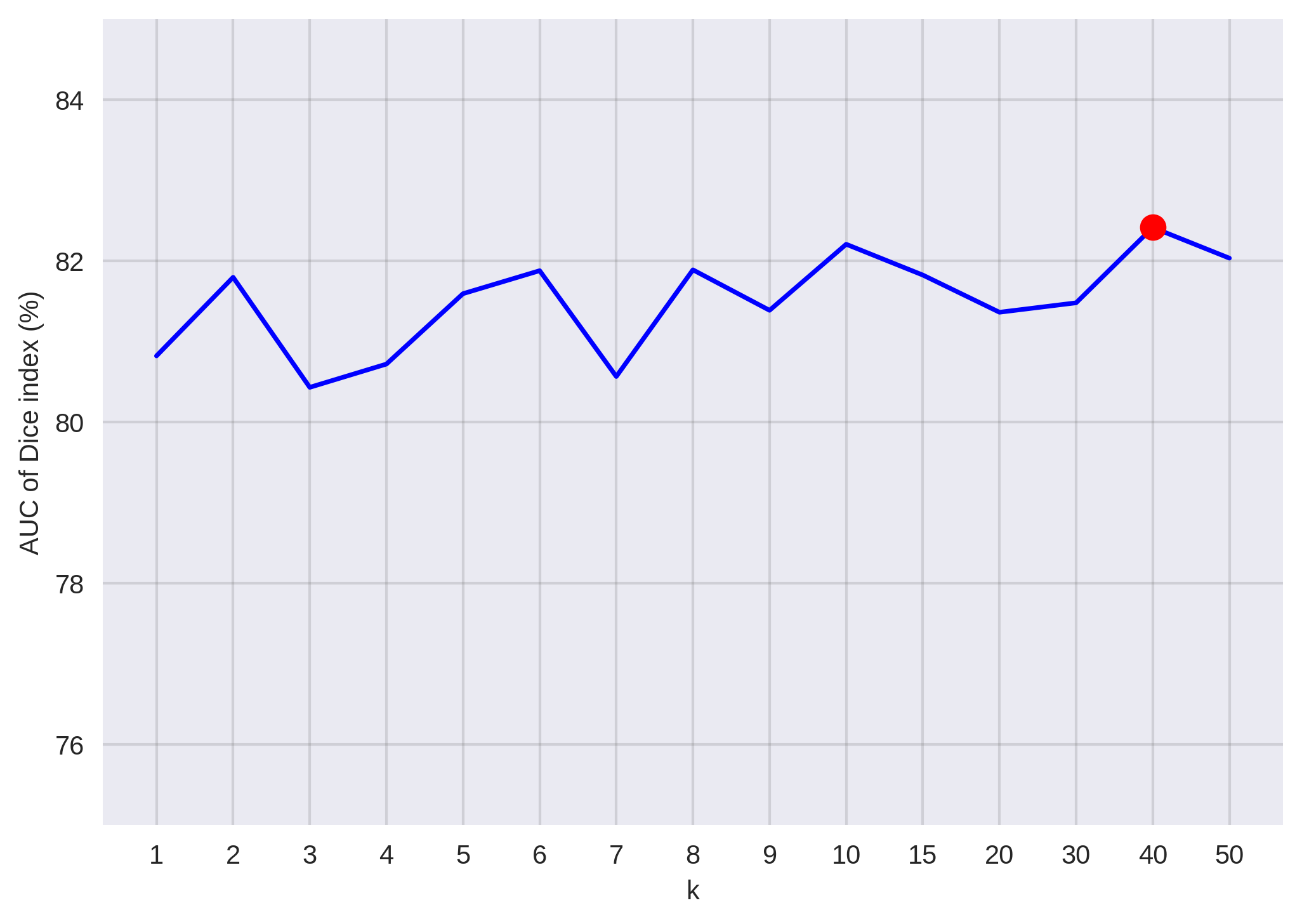}
  \caption{Ablation study over GlaS dataset (test set) over the hyper-parameter $k$ (x-axis). y-axis: AUC of Dice index (\%) of \textbf{25 queries for one trial}.
  AUC average $\pm$ standard deviation: ${81.49 \pm 0.59}$.
  Best performance in red dot: $k=40, AUC=82.41\%$.
  }
  \label{fig:abl-glas-knn}
\end{figure}

\begin{figure}[h!]
\centering
  \centering
  \includegraphics[scale=0.5]{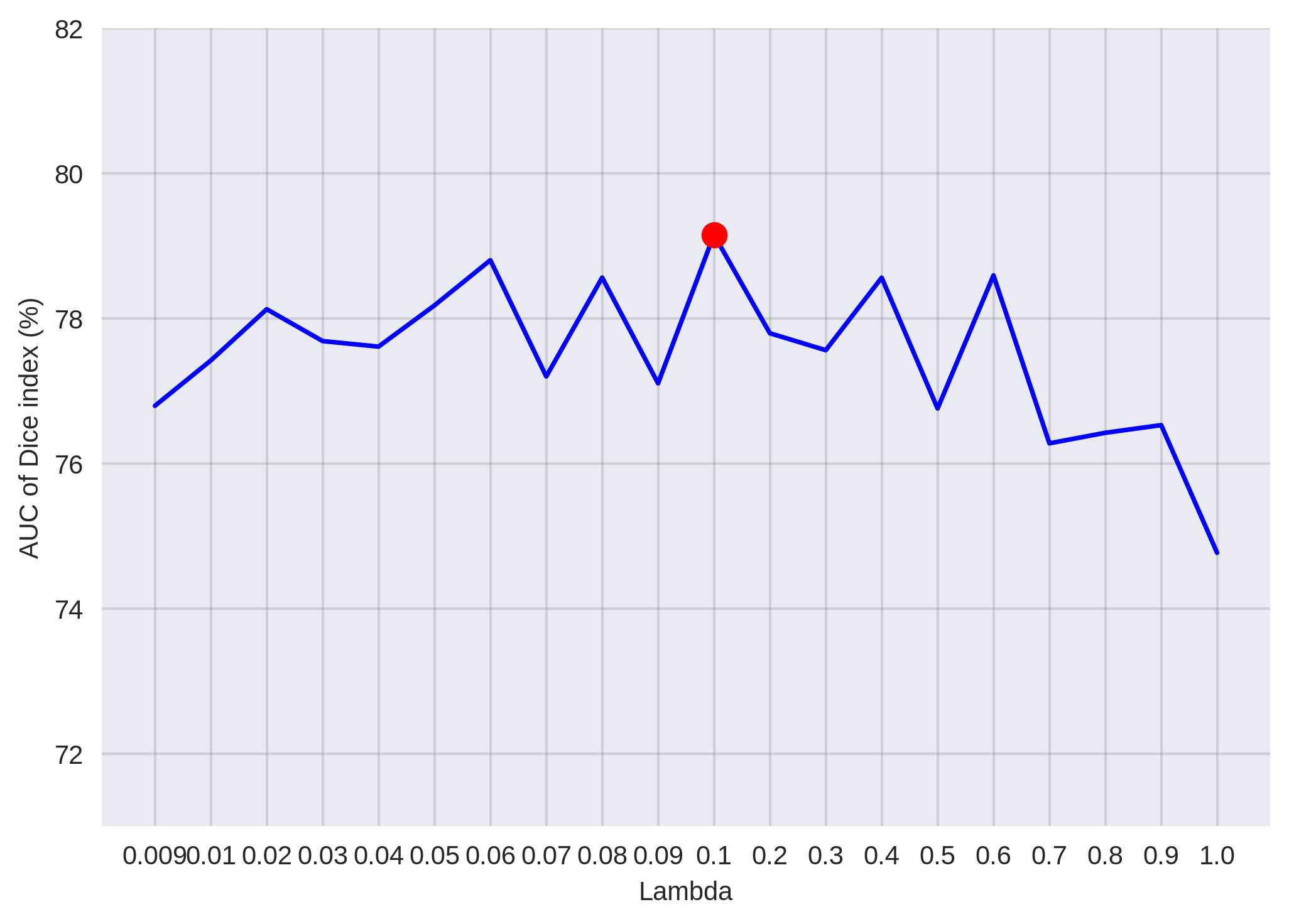}
  \caption{Ablation study over GlaS dataset (test set) over the hyper-parameter ${\lambda}$ (x-axis). y-axis: AUC of Dice index (\%) of \textbf{15 queries for one trial}.
  Best performance in red dot: $\lambda=0.1, AUC=79.15\%$.
  }
  \label{fig:abl-glas-lambda}
\end{figure}

\begin{figure}[h!]
\centering
  \centering
  \includegraphics[scale=0.5]{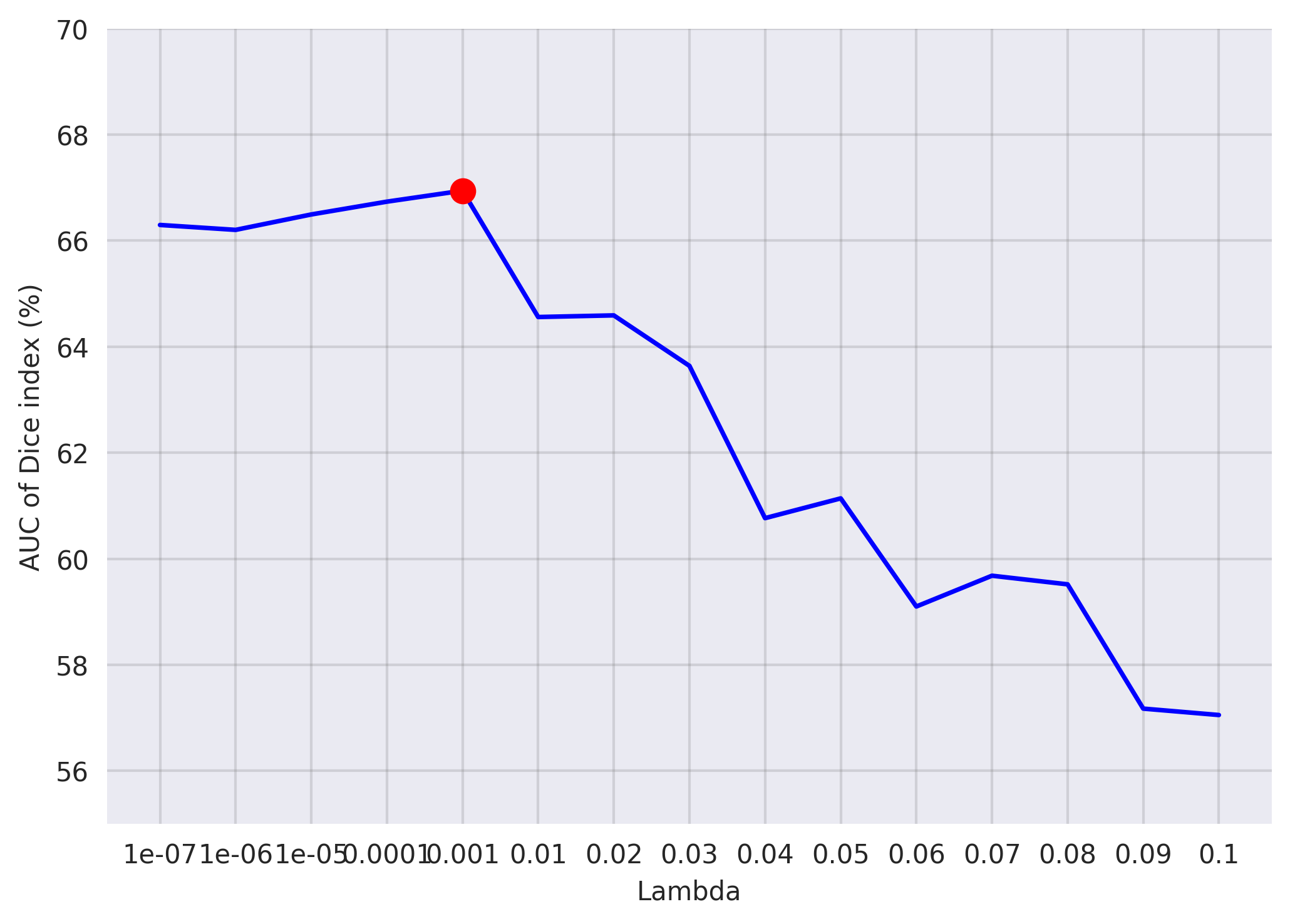}
  \caption{Ablation study over CUB dataset (test set) over the hyper-parameter ${\lambda}$ (x-axis). y-axis: AUC of Dice index (\%) of \textbf{5 queries for one trial}.
  Best performance in red dot: $\lambda=0.001, AUC=66.94\%$.
  }
  \label{fig:abl-cub-lambda}
\end{figure}

% ===================================================================================================
%                                           ABLATION STUDY
% ===================================================================================================

\subsection{Ablation study}
\label{sub:ablation}

We study the impact of $k$ and ${\lambda}$ on our method. Results are presented in Fig.\ref{fig:abl-glas-knn}, \ref{fig:abl-glas-lambda} for GlaS over ${k, \lambda}$; and in Fig.\ref{fig:abl-cub-lambda} for CUB over ${\lambda}$. Due to the expensive computation time required to perform AL experiments, we limited the experiments (${k, \lambda}$, number of trials, and \texttt{maxr}).
The obtained results of this study show that our method is less sensitive to ${k}$ (standard deviation of ${0.59}$ in Fig.\ref{fig:abl-glas-knn}).
In other hand, the method shows sensitivity to ${\lambda}$ as expected from penalty-based methods \citep{bertsekas1999nonlinear}.
However, the method seems more sensitive to ${\lambda}$ in the case of CUB than GlaS. CUb dataset is more challenging leading to more potential erroneous pseudo-annotation. Using Large ${\lambda}$ will systematically push the model to learn on the wrong annotation (Fig.\ref{fig:abl-cub-lambda}) which leads to poor results. In the other hand, GlaS seems to allow obtaining good segmentation where using large values of ${\lambda}$ did not hinder the performance quickly (\ref{fig:abl-glas-lambda}).
The obtained results recommend using small values that lead to better and stable performance. Using high values, combined with the pseudo-annotation errors, push the network to learn erroneous annotation leading to overall poor performance.

% ===================================================================================================
%                                           SIMILARITY MEASURE
% ===================================================================================================
\subsection{Similarity measure}
\label{subsec:similarity}
\begin{figure*}[hbt!]
\centering
  \centering
  \includegraphics[width=1.\linewidth]{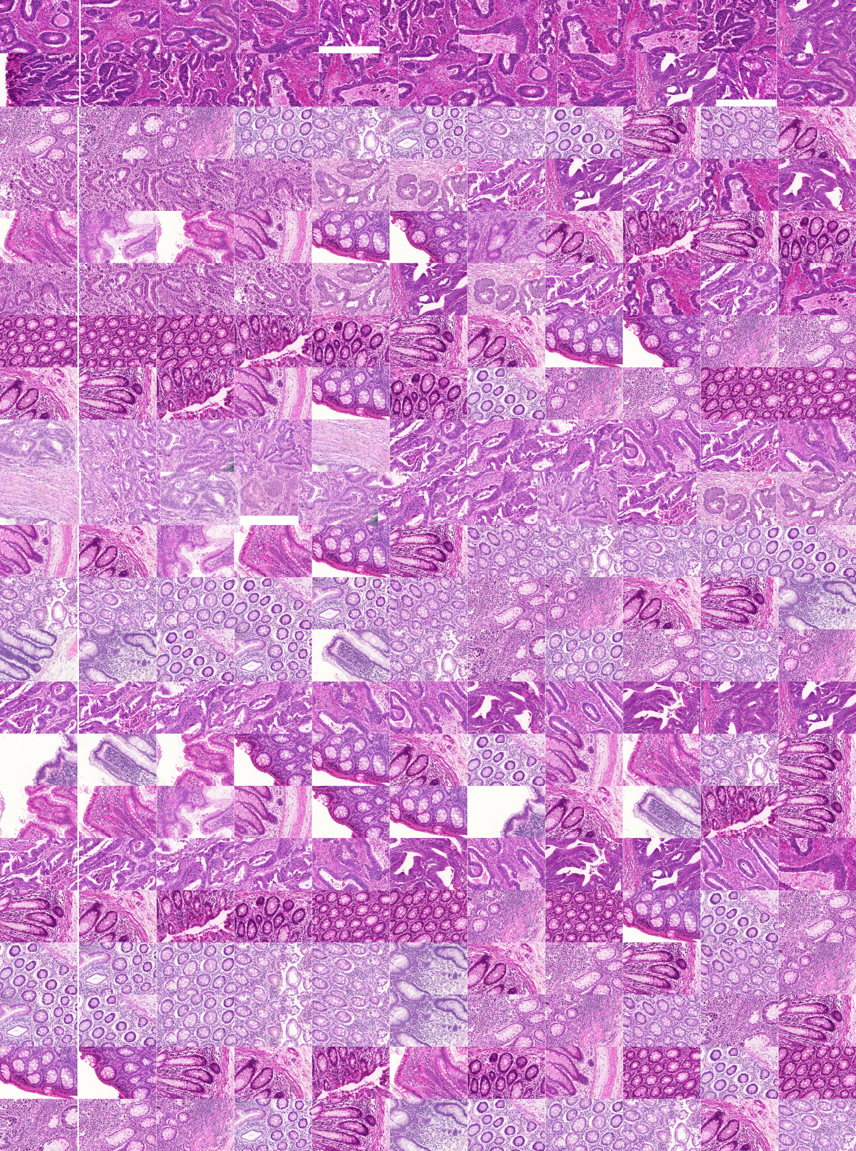}
  \caption{Examples of ${k}$-nn over GlaS dataset. The images represents the 10 nearest images to the first image in the extreme left ordered from the nearest.}
  \label{fig:glas-sim}
\end{figure*}

\begin{figure*}[hbt!]
\centering
  \centering
  \includegraphics[width=0.9\linewidth]{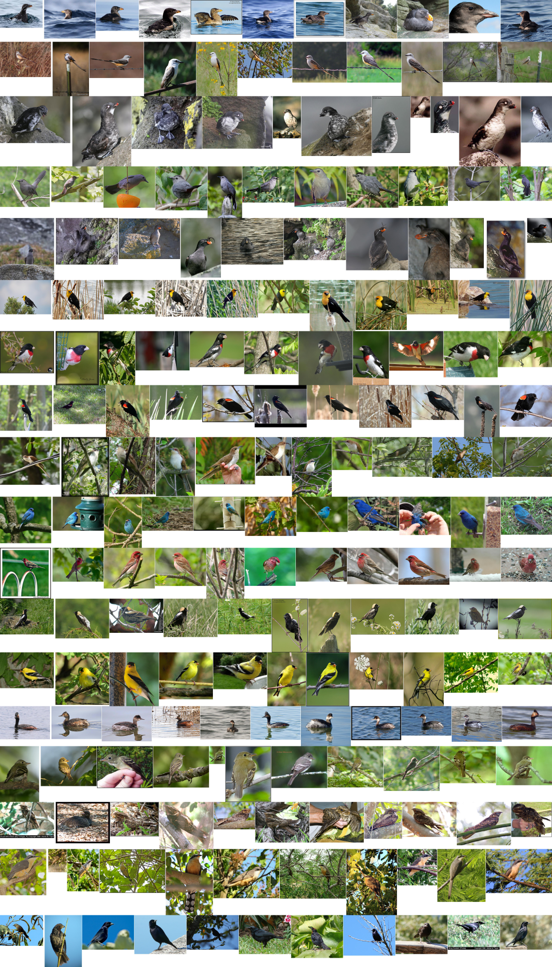}
  \caption{Examples of ${k}$-nn over CUB dataset. The images represents the 10 nearest images to the first image in the extreme left ordered from the nearest.}
  \label{fig:cub-sim}
\end{figure*}

In this section, we present some samples with their nearest neighbors. Although, it is difficult to quantitatively evaluate the quality of such measure.
Fig.\ref{fig:glas-sim} shows the case of GlaS. Overall, the similarity shows good behavior of capturing the general stain of the image which is what was intended for since the structure of such histology images is subject to high variation. Since the stain variation is one of the challenging aspects in histology images \citep{rony2019weak-loc-histo-survey}, labeling a sample with a common stain can help the model in segmenting other samples with similar stain.
The case of CUB, presented in Fig.\ref{fig:cub-sim}, is more difficult to judge the quality since the images contain always the same species within their natural habitat. Often, the similarity succeeds to capture the overall color, background which can help segmenting the object in the neighbors and also the background. In some cases, the similarity captures samples with large zoom-in where the bird color dominate the image.

\subsection{Predicted mask visualization}
\label{subsec:mask-vis}
\begin{figure*}[h!]
  \centering
  \includegraphics[width=0.95\linewidth]{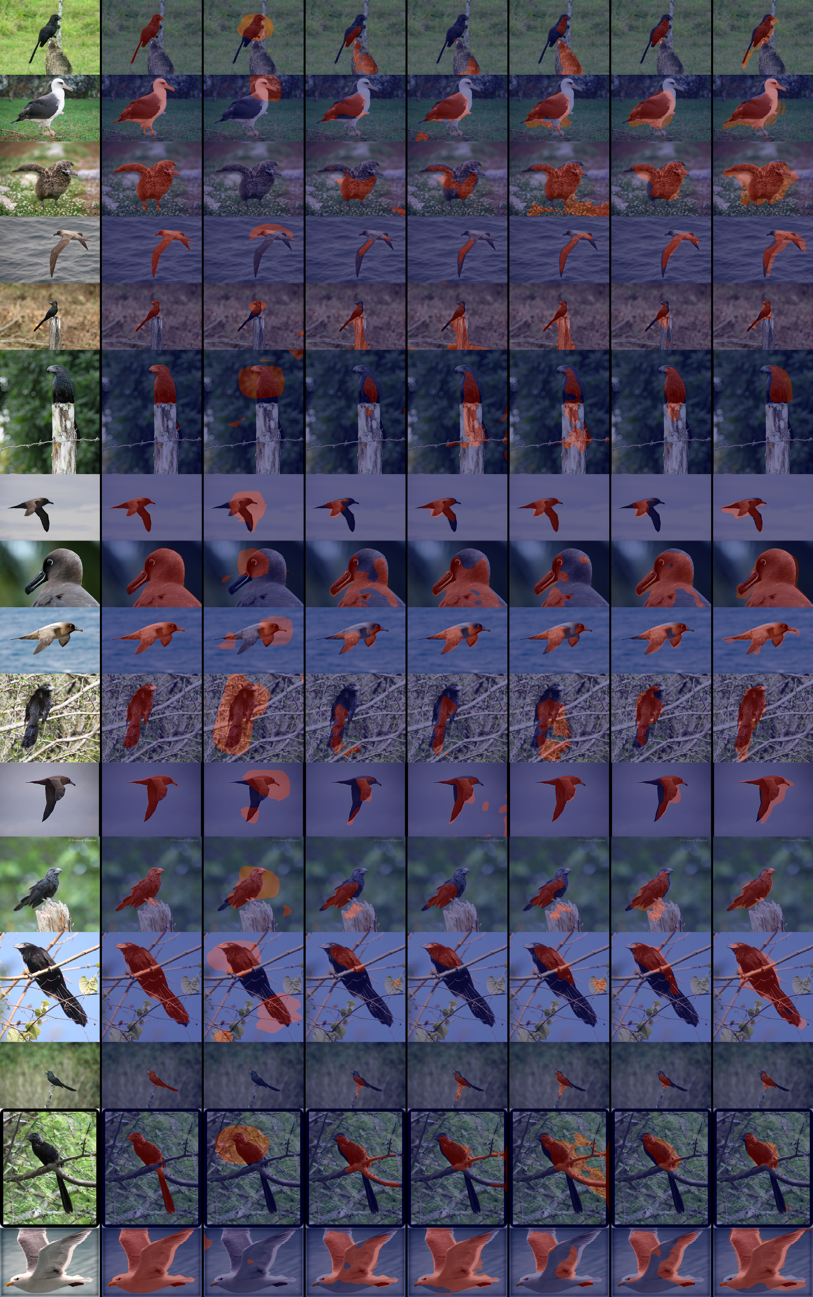} \\
  \includegraphics[width=0.95\linewidth]{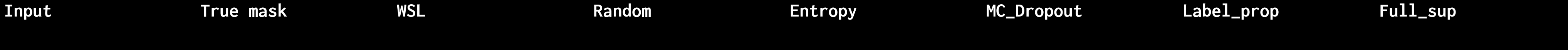}
  \caption{Qualitative results (on several CUB test images) of the predicted binary mask for each method after being trained in the first round ${r=1}$ (\ie after labeling 1 sample per class) using seed=0.  The average Dice index over the test set of each method is: ${40.16\%}$ (WSL), ${55.32\%}$ (Random), ${55.41\%}$ (Entropy), ${55.52\%}$ (MC\_Dropout), ${59.00\%}$ (Label\_prop), and ${75.29\%}$ (Full\_sup). (Best visualized in color.)}
  \label{fig:cub-results-visu}
\end{figure*}

Fig.\ref{fig:cub-results-visu} shows several test examples of predicted masks of different methods over CUB test set at the first AL round (${r=1}$) where only one sample per class has been labeled by the oracle. This interesting functioning point shows that by labeling only one sample per class, the performance of the average Dice index can go from ${39.08 \pm 08}$ for WSL method up to ${62.58 \pm 2.15}$ for Label\_prop and other AL methods. The figure shows that WSL tend to spot small part of the object in addition to the background leading high false positive. Using few supervision in combination with the proposed architecture, better segmentation is achieved by spotting large part of the object with less confusion with the background.

\FloatBarrier

% =============================================================================================
%                                       BIBLIOGRAPHY
% =============================================================================================

\bibliographystyle{apalike}
\bibliography{biblio}

\end{document}